\documentclass[runningheads]{llncs}
\usepackage[symbol]{footmisc}
\usepackage{graphicx}
\usepackage{amsmath,amssymb,amsfonts}
\usepackage{multirow}
\usepackage{algorithm}
\usepackage{algorithmicx}
\usepackage{algpseudocode}
\usepackage{hyperref}

\begin{document}
\title{SAM-kNN Regressor for Online Learning in Water Distribution Networks\thanks{We gratefully acknowledge funding from the VW-Foundation for the project \textit{IMPACT} funded in the frame of the funding line \textit{AI and its Implications for Future Society}, and funding from from the European Research Council (ERC) under the ERC Synergy Grant Water-Futures (Grant agreement No. 951424).}}
%
%
\author{Jonathan Jakob\inst{1}$^\dagger$ \and
Andr\'e Artelt \inst{1,3}$^\dagger$ \and
Martina Hasenj\"ager \inst{2} \and
Barbara Hammer \inst{1}}

\footnotetext{$^\dagger$First two authors contributed equally.}

\authorrunning{J. Jakob et al.}
%
\institute{Bielefeld University, Bielefeld, Germany \and
Honda Research Institute, Offenbach, Germany \and
University of Cyprus, Nicosia, Cyprus}
\maketitle              
\begin{abstract}
Water distribution networks are a key component of modern infrastructure for housing and industry. They transport and distribute water via widely branched networks from sources to consumers. 
In order to guarantee a working network at all times, the water supply company continuously monitors the network and takes actions when necessary -- e.g. reacting to leakages, sensor faults and drops in water quality. Since real world networks are too large and complex to be monitored by a human, algorithmic monitoring systems have been developed. A popular type of such systems are residual based anomaly detection systems that can detect events such as leakages and sensor faults. For a continuous high quality monitoring, it is necessary for these systems to adapt to changed demands and presence of various anomalies.

In this work, we propose an adaption of the incremental SAM-kNN classifier for regression to build a residual based anomaly detection system for water distribution networks that is able to adapt to any kind of change.

\end{abstract}
\section{Introduction}
Water is the foundation of (our) life -- we need water for drinking, cooking, hygiene and farming.
Water distribution networks (WDNs), which distribute water from the supplier to the customers, are therefore considered as critical infrastructure.
A major problem for water utility companies (and society in general) are anomalies that cause loss or contamination of water -- e.g. leakages such as pipe bursts, sensor faults, pollution, cyber-physical attacks, etc.~\cite{farley2003losses,alexander2019contamination,nikolopoulos2020cyber}.
Because of water shortages, among others caused by climate change, (drinking) water becomes an increasingly valuable resource that should not be wasted.
However, it was estimated that leakages in WDNs lead to a loss of more than 45 million $m^3$ of drinking water in developing countries -- even high developed countries such as the island of Cyprus looses approx. up to 25\% of their drinking water due to leaky pipes~\cite{liemberger2006challenge}.

Because of the increasing availability of sensors (e.g. pressure sensors) in WDNs, water utility companies nowadays use computer systems for (autonomously) monitoring their networks~\cite{makropoulos2019urban}.
These systems are realized using methods from engineering, statistics and machine learning (ML)~\cite{chan2018review}. While many different and successful methods for anomaly detection and localization have been proposed~\cite{wu2017review}, these methods are usually not able to adapt to an occurring change or anomaly.
These systems might be able to detect anomalies but once the anomaly is detected, they are ``blind'' for everything else that happens while the detected anomaly is present -- the systems must be recalibrated or refitted which becomes challenging because a large amount of data (and therefore collection time) is needed.
Adaptation to changes -- in case of WDNs anomalies or simply changes in the water consumption behavior of the customers (i.e. changed demand) -- can be natural handled by online learning methods.

Online learning~\cite{gama2012survey} can be considered as a sub-field of machine learning which deals with models that are trained incrementally -- i.e. they can learn from a data stream instead of a fixed training set only. For example, they can be used for electricity price prediction \cite{onlinereg1} or electric load forecasting \cite{onlinereg2}.

In this work, we contribute to online learning for regression problems and water distribution networks as a particular field of application. More specifically, our contributions are:
\begin{itemize}
    \item We propose SAM-kNN regression, a memory based online learner for regression problems.
    \item We evaluate our proposed SAM-kNN regression method in the context of anomaly detection in water distribution networks.
\end{itemize}

The remainder of this paper is structured as follows: After briefly reviewing related work in Section~\ref{sec:relatedwork}, we introduce the problem setting we are considering in this work (see Section~\ref{sec:problemsetting}). Next, in Section~\ref{sec:samknn} we propose SAM-kNN regression for online learning, which we empirically evaluate in the context of anomaly detection in water distribution networks (see Section~\ref{sec:experiments}). Finally, this work closes with a summary and conclusion in Section~\ref{sec:conclusion}.

\section{Related Work}\label{sec:relatedwork}
Incremental or online learning is a machine learning paradigm in which a model is updated after each data sample that is fed into it. This paradigm is especially suited for large data sets, that are too big to be processed in batch fashion, or in situations where data becomes available only sample after sample -- e.g. in the form of a  potentially infinite streams of data.
Incremental learning has made great strides in recent years \cite{online1}, with most applications set in a classification environment \cite{online2,online3,online4,online5,online6,online7}. 

One particular system is the SAM-kNN classifier \cite{sam}. This incremental algorithm was created to perform on long data streams, using its internal memory structure to alleviate the problem of catastrophic forgetting when frequent concept changes are expected.

On the other hand, only a few approaches that utilize incremental learning for regression problems exist. In \cite{onlinereg2} the authors use an incremental variant of the support vector regressor (SVR) to build models for electricity price prediction. \cite{onlinereg1} uses a similar SVR, paired with phase space reconstruction for time series to facilitate electrical load forecasting. In \cite{onlinereg3} a wide range of incremental regression algorithms are compared for their use in exoskeleton control.

However, all of these publications utilize standard incremental algorithms and do not explicitly build a new model to work with. We, on the other hand, propose a reformulation of the SAM-kNN classifier for regression as a standalone algorithm.

\section{Problem Setting}\label{sec:problemsetting}

Incremental or online regression is the task of predicting a response variable $y \in \mathbb{R}$ from a stream $S = \{x_1, x_2, x_3, ...\} $ of variables $X \in \mathbb{R}^n$. Hereby, a new instance of the incremental model is learned for each incoming sample of the data stream.

We work on water distribution networks that have several internal nodes $n$. These nodes are equipped with sensors, that measure water pressure and flow rate. Each sensor provides read outs at specific time intervals $t$. For every node, this creates a potentially infinite data stream of sensors values $S = \{(s_1, y_1), (s_2, y_2), (s_3, y_3), ...\}$, where $s_i \in \mathbb{R}^{n-1}$ represents the sensor values of all but one node in the network and the predictor variable $y_i \in \mathbb{R}$ represents the sensor value at the remaining node.

This means, that we use the read outs of $n-1$ nodes to predict the value of the $n$th node using an incremental regression algorithm. Said algorithm processes the stream $S$ instance after instance by generating a sequence of models $H = \{h_1, h_2, h_3, ...\}$, where $h_{i-1}(s_i) = \hat{y}_i$. After each prediction the true value $y_i$ is revealed and a new model $h_{i}$ is learned.

We use the \emph{Interleaved train test error (ITTE)} as a cost function to train the model:

\begin{equation}
    E(S) = \sqrt{\frac{1}{t} \sum^t_{i=1} (h_{i-1}(s_{i}) - y_{i})^2} 
\end{equation}

\noindent
This ITTE measures the \emph{Root Mean Squared Error (RMSE)} over every model $h_i$ up to a given time point $t$.

Whenever the local error $h_{i-1}(s_i) - y_i$ exceeds a certain threshold, this means, that more water than predicted flows through the observed node. This is taken as an indication for a water leak and an alarm will be triggered. Being an incremental algorithm, our model will than automatically adjust to the new circumstances so that accurate prediction of the water flow will be maintained throughout the leak. Our model, which will be explained in detail in the next section, has the capability to remember long term concepts and therefore, as soon as the water leak is fixed, it will revert back to the normal circumstances.

\section{Model}\label{sec:samknn}
Our proposed model is an adaption of the Self Adjusting Memory (SAM) \cite{sam}, an incremental classifier, to regression. This approach is based on two distinct internal memories, the Short-Term (STM) and the Long-Term memory (LTM). Hereby, the STM is a dynamic sliding window over the last $m$ samples, that is supposed to only hold the most recent concept of the data stream:

\begin{equation}
    M_{ST} = \{(x_i, y_i) \in \mathbb{R}^n \times \mathbb{R} \ | \ i = t-m+1, ... , t\}
\end{equation}

\noindent
The LTM, on the other hand, is a collection of $p$ samples, which hold older concepts, that do not contradict the STM and might still be of use in the future:

\begin{equation}
    M_{LT} = \{(x_i, y_i) \in \mathbb{R}^n \times \mathbb{R} \ | \ i = 1, ..., p\}
\end{equation}

\noindent
Additionally, there is the combined memory (CM), which is a simple union of the STM and the LTM:

\begin{equation}
    M_{C} = M_{ST} \cup M_{LT}
\end{equation}

\noindent
Each memory induces a kNN regressor that can be used independently from the others. To determine which kNN is used for every new incoming data sample, the ITTE (see section \ref{sec:problemsetting}) is tracked for all sub-models and the one with the lowest current ITTE is chosen.

\subsection{Model Parameters}
The proposed model has three parameters that are continuously adapted during deployment:

\begin{enumerate}
    \item The size $m$ of the STM sliding window
    \item The data samples in the LTM
    \item The ITTEs of the three sub-models
\end{enumerate}

\noindent
Additionally, there are three hyperparameters that can be chosen robustly and are set before deployment:

\begin{enumerate}
    \item The number of neighbours $k$
    \item The minimum size $L_{min}$ of the STM 
    \item The maximum size $L_{max}$ of the LTM 
\end{enumerate}

\subsection{Model Adaption}
Whenever a new data sample arrives, it is added to the STM, which means that this memory grows continuously. However, since it is supposed to hold only the most recent concept, a reduction of the STM window size is performed on a regular basis. This is facilitated by testing smaller window sizes at every iteration and choosing the one that is optimizing the ITTE. Tested windows are:

\begin{equation}
    M_l = \{(x_{t-l+1}, y_{t-l+1}), ..., (x_t, y_t)\} 
\end{equation}

\noindent
where $l \in \{m, m/2, m/4 ...\}$ and $l \geq L_{min}$.

\begin{equation}
    M_{ST_{t+1}} =  \underset{S \in \{M_m, M_{m/2}, ...\}}{\mathrm{argmin}} E(S)
\end{equation}

\noindent
Whenever the STM is shrunk in size, the data samples $O_t$ that fall out of the sliding window are not discarded. 

\begin{equation}
    O_t = M_{ST_t} \setminus M_{ST_{t+1}}
\end{equation}

\noindent
Instead, they undergo a cleaning process, and those, that are still consistent with the new STM are added to the LTM. Afterwards, the whole of the LTM is cleaned as well, to ensure consistency with the STM at all times. When the LTM reaches its maximum size, samples get discarded in a way that ensures minimal information loss.

\subsection{Cleaning Process}
The process to clean a set of samples with respect to the STM is defined in the following way: 

\noindent
A set $A$ is cleaned by another set $B$ regarding an example $(x_i, y_i) \in B$

\begin{equation}
    clean : (A, B, (x_i, y_i)) \mapsto \hat{A}
\end{equation}

\noindent
where $A, B, \hat{A} \subset \mathbb{R}^n \times \mathbb{R}$ and $(x_i, y_i) \in B$. \\

\noindent
$\hat{A}$ is defined in five steps:

\begin{enumerate}
    \item Determine the $k$ nearest neighbours of $x_i$ in $B \setminus (x_i, y_i)$ and find the maximum distance
    
        \begin{equation}
            \Delta_x^\ast = \max\big\{d(x_i,x) \ | \ x \in N_k(x_i, B \setminus (x_i, y_i))\big\}
        \end{equation}
        
    \item Compute the maximum weighted difference of $y_i$ and $y \in N_k(x_i, B \setminus (x_i, y_i))$
    
        \begin{equation}
            \Delta_y^\ast = \max \Bigg\{ \Bigg(\frac{y_i - y}{e^\frac{x_i - x}{\Delta_x^\ast}} \Bigg) \ | \ y \in N_k(x_i, B \setminus (x_i, y_i)) \Bigg\}
        \end{equation}
    
    \item Determine all samples in $A$ that are within $\Delta_x^\ast$ of $x_i$
    
        \begin{equation}
            C = \big\{(x, y) \in A \ | \ d(x_i, x) < \Delta_x^\ast\big\}
        \end{equation}
        
    \item Compute the weighted differences of $y_i$ and $y \in C$
    
        \begin{equation}
            \Delta_y = \Bigg\{ \Bigg(\frac{y_i - y}{e^\frac{x_i - x}{\Delta_x^\ast}} \Bigg) \ | \ y \in C\Bigg\}
        \end{equation}
        
    \item Discard samples from $C$ that have a larger weighted difference than $\Delta_y^\ast$
    
        \begin{equation}
            \hat{A} = A \setminus \{(x,y) \in C \ | \ \Delta_y(x) > \Delta_y^\ast\}
        \end{equation}
    
\end{enumerate}

\noindent
Furthermore, the cleaning operation for the full set $B$

\begin{equation}
    clean : (A, B) \mapsto \hat{A}_{|B|}
\end{equation}

\noindent is defined by iteratively applying the former cleaning for all $(x_i, y_i) \in B$

\begin{align*}
    \hat{A}_0 &= A \\
    \hat{A}_{t+1} &= clean(\hat{A}_t, B, (x_{t+1}, y_{t+1}))
\end{align*}

\noindent
In summary, when the STM is shrunk in size, the process to clean the discarded set $O_t$ is described by the operation:

\begin{equation}
    clean(O_t, M_{ST_{t+1}})
\end{equation}

\noindent
After that, the LTM is cleaned as well:

\begin{equation}
    clean(M_{LT_t}, M_{ST_{t+1}})
\end{equation}

\subsection{Compression of the LTM}
When new samples are added to the LTM while it reaches maximum capacity, old samples need to be discarded. To avoid a significant information loss, samples are discarded in an iterative process one after another until $|M_{LT}| < L_{max}$ again. Hereby, the data sample with the lowest distance to any other two samples is chosen for every iteration of the discarding process. 

\subsection{Final Model}
The complete pseudocode of our proposed model is given in Algorithm \ref{algo:sam}.
\begin{algorithm}[H]
\caption{SAM-kNN Regression}\label{algo:sam}
\textbf{Input:} Data stream $S$, one $s_i$ at a time \\
\textbf{Output:} $\hat{y}_i$ for every $s_i$
\begin{algorithmic}[1]
 \State $M_{ST}$, $M_{LT}$ = $\{s_0, .., s_{L_{min}}\}$ \Comment{Initialize STM and LTM}
 \State $E_{ST}, E_{LT}, E_{C} = 0$ \Comment{Initialize tracked errors}
  \For{$s_i \in S \setminus \{s_0, .., s_{L_{min}}\}$} \Comment{Loop over the remaining data stream}
 	\State $BM = argmin (E_{ST}, E_{LT}, E_{C})$ \Comment{Find best memory with lowest error}
 	\State $\hat{y_i} = kNN_{BM}(s_i)$ \Comment{Predict with kNN of best memory}
 	\State Update $E_{ST}, E_{LT}, E_{C}$
 	\State $M_{ST} = M_{ST} \ \cup \ \{s_i\} $ \Comment{Add current sample to STM}
 	\State Evaluate smaller STM sizes
 	\If{STM is reduced} 
 	    \State $O_t = M_{ST_t} \setminus M_{ST_{t+1}}$ \Comment{Take discarded samples from STM}
 		\State $clean(O_t, M_{ST_{t+1}})$ \Comment{Clean discarded samples with respect to new STM}
 		\State $M_{LT} = M_{LT} \ \cup \ clean(O_t, M_{ST_{t+1}})$ \Comment{Add cleaned samples to LTM}
 		\State $clean(M_{LT}, M_{ST_{t+1}})$ \Comment{Clean new LTM with respect to new STM}
 	\EndIf
 \EndFor
\end{algorithmic}
\end{algorithm}

\section{Experiments}\label{sec:experiments}
We empirically evaluate our proposed method in an online scenario for detecting leakages and sensor faults in water distribution networks -- all experiments are implemented in Python\footnote{Implementation is available on GitHub: \url{https://github.com/andreArtelt/SAM-kNN-Regressor\_OnlineLearning\_WDNs}}.

\subsection{Data}
\begin{figure}
    \centering
    \includegraphics[width=\textwidth]{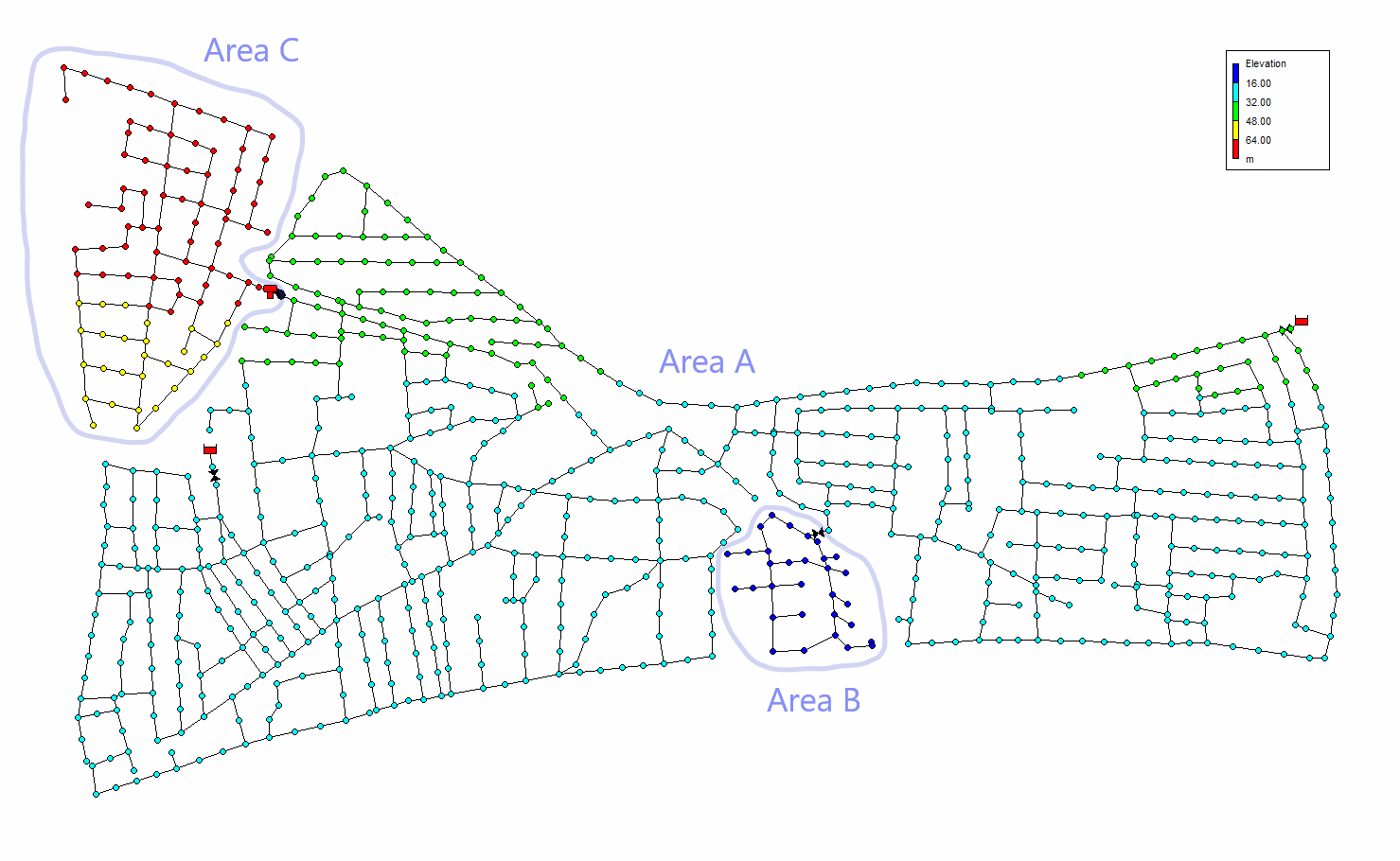}
    \caption{L-Town network~\cite{vrachimis2020battledim} -- we only use ``Area A'' where we have $29$ pressure sensors.}
    \label{fig:l_town}
\end{figure}
We use a version of the L-Town water distribution network as used in~\cite{vrachimis2020battledim}, as a prominent realistic benchmark for anomaly detection -- we use Area A only which consists of $661$ nodes, $766$ links, and $29$ (optimally placed) pressure sensors. We build and simulate $10$ scenarios where the first $5$ scenarios each contain a single leakage -- we vary position, time and size of the leakage -- and the remaining $5$ scenarios each contain a single different sensor fault (position is varied):
\begin{itemize}
    \item Scenario 6: Sensor measurement is overflowing over time -- i.e. it is going to infinity over time.
    \item Scenario 7: Gaussian noise is added to the sensor measurement.
    \item Scenario 8: A constant value is added to the sensor measurement.
    \item Scenario 9: Sensor measurement is set equal to zero.
    \item Scenario 10: Sensor measurement is shifted by a small amount.
\end{itemize}
Each scenario is simulated (using WNTR~\cite{wntr}) for $3$ month and pressure sensors are sampled every $5$ minutes.

The data stream of sensor measurements is post processed by using a sliding window of size $4$ -- we average all samples dimensional wise in this time window, so that we end up with $29$ dimensional samples.

\subsection{Setup}
We compare our proposed SAM-kNN regressor (see Section~\ref{sec:samknn}) to several other online learning regressors. In order to justify the introduced overhead of the SAM architecture, we compare its performance to vanilla regressors (kNN regression and linear regression) wrapped as online learners by using the river toolbox\footnote{\url{https://github.com/online-ml/river}}.

For each pressure sensor, we build a corresponding virtual sensor based on all other pressure sensors -- i.e. we try to predict (using a regressor) the pressure based on the past pressure values of all other pressure sensors (see Section~\ref{sec:problemsetting}). These virtual sensors are then used for a residual based anomaly detection -- i.e. an alarm (detected anomaly) is raised when the predicted pressure value deviates to much from the observed pressure measurement.
For each scenario, the processed data stream is fed as batches of $200$ samples to the regressors (realizing the virtual sensors).

For each regressor, we evaluate the performance of the resulting anomaly detector -- since we are interested in the detection of a single anomaly (leakage and sensor fault), we report true positives (TP), false positives (FP) and false negatives (FN). Note that the true positives and false negatives are always either $0$ or $1$ because we only check whether an alarm was raised when the single anomaly was present or not -- however, for the false positives, we count every single false alarm.

\subsection{Results}
The results for leakage detection, for each regressor and each scenario, are shown in Table~\ref{tab:exp_results:leakage}. Likewise, the results for sensor fault detection are shown in Table~\ref{tab:exp_results:sensorfault}.
\begin{table}[]
    \centering
    \caption{Leakages: Evaluation of residual based anomaly detection in water distribution networks -- note that each scenario consists of approx. $23000$ samples.}
    \begin{tabular}{|c|c|c|c|c|c|c|}
        \hline
         Method & Metric & Scenario 1 & Scenario 2 & Scenario 3 & Scenario 4 & Scenario 5 \\
         \hline
         \multirow{3}{*}{SAM-kNN}
         & TP & $1$ & $1$ & $1$ & $1$ & $1$ \\ 
         & FP & $48$ & $20$ & $3$ & $20$ & $17$ \\ 
         & FN & $0$ & $0$ & $0$ & $0$ & $0$ \\ 
         \hline
         \multirow{3}{*}{kNN}
         & TP & $1$ & $1$ & $1$ & $1$ & $1$ \\ 
         & FP & $17057$ & $19216$ & $11146$ & $19082$ & $18751$ \\ 
         & FN & $0$ & $0$ & $0$ & $0$ & $0$ \\ 
         \hline
         \multirow{3}{*}{Linear regression}
         & TP & $0$ & $0$ & $0$ & $0$ & $0$ \\ 
         & FP & $0$ & $0$ & $0$ & $0$ & $0$ \\ 
         & FN & $1$ & $1$ & $1$ & $1$ & $1$ \\ 
         \hline
    \end{tabular}
    \label{tab:exp_results:leakage}
\end{table}
\begin{table}[]
    \centering
    \caption{Sensor faults: Evaluation of residual based anomaly detection in water distribution networks -- note that each scenario consists of approx. $23000$ samples.}
    \begin{tabular}{|c|c|c|c|c|c|c|}
        \hline
         Method & Metric & Scenario 6 & Scenario 7 & Scenario 8 & Scenario 9 & Scenario 10 \\
         \hline
         \multirow{3}{*}{SAM-kNN}
         & TP & $1$ & $1$ & $1$ & $1$ & $1$ \\ 
         & FP & $155$ & $20$ & $157$ & $96$ & $156$ \\ 
         & FN & $0$ & $0$ & $0$ & $0$ & $0$ \\ 
         \hline
         \multirow{3}{*}{kNN}
         & TP & $1$ & $1$ & $1$ & $1$ & $1$ \\ 
         & FP & $18596$ & $18596$ & $18596$ & $18596$ & $18596$ \\ 
         & FN & $0$ & $0$ & $0$ & $0$ & $0$ \\ 
         \hline
         \multirow{3}{*}{Linear regression}
         & TP & $1$ & $0$ & $0$ & $0$ & $0$ \\ 
         & FP & $0$ & $0$ & $0$ & $0$ & $0$ \\ 
         & FN & $0$ & $1$ & $1$ & $1$ & $1$ \\ 
         \hline
    \end{tabular}
    \label{tab:exp_results:sensorfault}
\end{table}
We observe the same/similar effects for both types of anomalies (leakages and sensor faults):
We observe that linear regression completely fails to detect the anomaly -- this indicates that a linear model is not sufficient to model the hydraulic dynamics in the water distribution network and hence fails to detect any anomalies. There is only one exception: For scenario 6 the linear model is able to detect the sensor fault -- recall that in this particular scenario the sensor fault is characterized by a slowly overflowing sensor measurements (i.e. the pressure value goes to infinity), which is expected to be easily detected because it is a very ``loud'' fault.
The kNN model shows good performance in detecting the anomaly but has a huge number of false positives -- i.e. it is too sensitive and raises lots of false alarms.
Our proposed SAM-kNN shows the best performance -- it is able to consistently detect the anomalies while having a small numbers of false positives only. The huge reduction of the false positives in comparison to vanilla kNN indicates that the overhead introduced by the SAM architecture actually pays off.

\section{Summary \& Conclusion}\label{sec:conclusion}
Inspired by the SAM-kNN classifier, we proposed SAM-kNN regressor as an online learner for regression. In contrast to other online learners, our proposed method comes with a memory component which allows it to remember past concepts quite easily.
We empirically evaluated our proposed online learner in an anomaly detection scenario for a realistic water distribution network -- our proposed online learner consistently outperforms other standard online learners.

Although our proposed method shows good performance for leakage and sensor fault detection (two very common anomalies), it is unclear, how well it performs for other (more complex) types of anomalies such as cyber-physical attacks, etc.
Furthermore, another challenge is high-dimensional data -- in this work we had $29$ dimensional which is already somewhat high but can be still managed by our kNN based method. However, in case of really high dimensional data, kNN will encounter performance problems -- e.g some kind of integrated dimensionality reduction might be required.
We leave these aspect as future work.

\section*{Acknowledgment}
We acknowledge the bachelor thesis by Augustin Harter (University of Bielefeld) and Yannik Sander (University of Bielefeld) which served as a mental starting point for this work.

%
%
%
\bibliographystyle{splncs04}
\bibliography{mybibliography}

\end{document}